# Fast Design Space Exploration of Nonlinear Systems: Part I

Sanjai Narain, Emily Mak, Dana Chee, Brendan Englot, Kishore Pochiraju, Niraj K. Jha, Karthik Narayan

*Abstract*— System design tools are often only available as input-output blackboxes: for a given design as input they compute an output representing system behavior. Blackboxes are intended to be run in the forward direction. This paper presents a new method of solving the "inverse design problem" namely, given requirements or constraints on output, find an input that also optimizes an objective function. This problem is challenging for several reasons. First, blackboxes are not designed to be run in reverse. Second, inputs and outputs can be discrete and continuous. Third, finding designs concurrently satisfying a set of requirements is hard because designs satisfying individual requirements may conflict with each other. Fourth, blackbox evaluations can be expensive. Finally, evaluations can sometimes fail to produce an output due to non-convergence of underlying numerical algorithms. This paper presents CNMA, a new method of solving the inverse problem that overcomes these challenges. CNMA tries to sample only the part of the design space relevant to solving the inverse problem, leveraging the power of neural networks, Mixed Integer Linear Programs, and a new learning-from-failure feedback loop. The paper also presents a parallel version of CNMA that improves the efficiency and quality of solutions over the sequential version, and tries to steer it away from local optima. CNMA's performance is evaluated against conventional optimization methods for seven nonlinear design problems of 8 (two problems), 10, 15, 36 and 60 real-valued dimensions and one with 186 binary dimensions. Conventional methods evaluated are stable, off-the-shelf implementations of Bayesian Optimization with Gaussian Processes, Nelder Mead and Random Search. The first two do not produce a solution for problems that are high-dimensional, have both discrete and continuous variables or whose blackboxes fail to return values for some inputs. CNMA produces solutions for all problems. When conventional methods do produce solutions, CNMA improves upon their performance by up to 87%.

*Index Terms*— Blackbox optimization; constrained optimization; Mixed-Integer Linear Program (MILP); neural networks; optimization; sample efficiency.

## I. INTRODUCTION

System design knowledge is often encapsulated inside blackboxes such as simulators, spreadsheets and program scripts. Blackboxes are, typically, nonlinear functions that accept a design as input and produce a representation of system behavior as output. New designs can be created by solving the "inverse problem": from requirements or constraints on blackbox output, compute an input that also optimizes an objective function.

Relevant to solving this problem are the mature, constrained nonlinear optimization methods [1, 2, 3, 4, 5, 6, 7, 8, 9, 10].

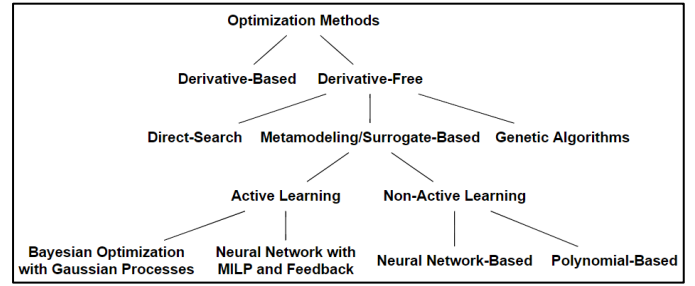

*Figure 1. Taxonomy of Nonlinear Optimization Methods*

Also available is a large online collection of these methods [11] along with a companion guide [12]. These methods can be adapted to solve the problem of finding $x$ that optimizes an objective function $\phi(x, F(x))$ subject to a constraint $P(x, F(x))$ where $F$ is a blackbox function, $\phi$ an objective function and $P$ a constraint or a requirement. Any solution $x$ is a solution to the inverse problem since $x$ is an input to $F$ that optimizes $\phi$ and satisfies $P$ on the output of $F$.

Figure 1 presents a taxonomy of these methods. These fall into two categories: derivative-based and derivative-free. The former compute derivatives of the objective function to determine the direction in which to search for a point where the derivative becomes zero. They are restricted to smooth, continuous functions. Thus, they do not apply to functions with discontinuities or discrete variables. Moreover, derivative computation is not sample-efficient in that it requires a large number of function evaluations. Thus, these methods are infeasible when function evaluation is expensive, as is often the case for blackboxes.

Derivative-free methods [5, 7] try to overcome the limitations of derivative-free ones. One class of such methods is called *direct search* whose well-known members include Nelder-Mead (NM) [6] and COBYLA [13]. They maintain a simplex (convex hull) of points around the current point and use it to compute the next point to sample in the direction of the optima. These methods require starting points whose incorrect choices can cause the methods to be stuck in local optima.

Another class of derivative-free methods is metamodeling or surrogate-based [7, 8]. They do not require starting points. Instead, they sample the blackbox function at some set of points and construct a surrogate model by fitting the values to a mathematical expression or by using machine learning. Active-learning metamodeling methods are conservative in the number of samples they evaluate. They do this by constructing a merit or acquisition function from the surrogate model. This merit function is optimized to compute the best point to sample next.

If this point does not satisfy a halting condition, the point is added to the set of samples and the search is restarted. Examples include ALAMO [13] and Bayesian Optimization (BO) with Gaussian Processes (BO/GP) [9]. In geostatistics, BO/GP is called Kriging [1] and the use of Kriging for circuit design is reported in [14]. In earlier BO/GP versions, the complexity of building surrogate functions was cubic in the number of samples [1] although asymptotically faster versions on GPUs are reported in [15]. To optimize with constraints, various extensions to BO/GP have been presented in [16, 17, 18, 19]. Parallel BO/GP algorithms have been presented in [20, 21, 22].

Non active-learning metamodeling methods, e.g., [23, 24, 25, 26], directly use the surrogate as a fast evaluator of the blackbox function for use by optimization methods, including derivative-based ones. They may not be feasible for higher-dimensional functions where the number of samples needed to construct an accurate enough surrogate may be astronomical.

Genetic algorithms [27] form a third type of derivative-free optimization. A population of samples is maintained that is systematically improved over multiple generations. While such algorithms can avoid local optima and operate on functions with discrete and continuous variables, they may converge to the final solutions slowly. The sister Part II of this paper [28] shows how the conjunction of genetic algorithm and CNMA can overcome this problem.

The above methods can be susceptible to failure of objective functions to evaluate, as can happen, for example due to non-convergence of computational fluid dynamics simulators [29]. If an artificial value has to be assigned to the function it could distort the simplexes and surrogates.

Many of the above methods handle constraints indirectly by reducing constrained optimization with a sequence of one or more unconstrained optimization problems [2, 4, 30, 31]. The principle is to encode the cost of violating a constraint as a penalty/barrier function and rely on an optimization engine to drive this cost to zero. For example, to model the constraint $x \leq 1$, the term $\max(0, x-1)$ can be added to the objective function. Whenever $x > 1$, the term would evaluate to a positive number. Hence, the optimization engine (doing minimization) would search in a region where the cost is zero. However, such functions can distort the shape of the new objective function making it harder to find the optima. Reducing such distortion requires substantial creativity on the part of the penalty/barrier function designer [4, 32].

This paper presents CNMA (it stands for Constrained optimization with Neural networks, MILP and Active learning), a new surrogate-based method for solving the inverse problem for blackboxes. Formally, CNMA finds values of $x, y$ that optimize $\phi(x, y)$ such that $F(x) = y \wedge P(x, y)$ where $F$ is a potentially nonlinear function available as a blackbox, $x, y$ are vectors of discrete and continuous variables, $\phi$ is a linear function and $P$ is a linear constraint. $P$ can also be a conjunction of several constraints. Note that $\wedge$ denotes logical conjunction (AND). This is a straightforward reformulation of the earlier inverse problem definition, with $y$ being an explicit handle on the output. This reformulation allows a natural implementation using the constituent technologies. *As shown in Section V.H, CNMA handles nonlinear objective functions and constraints by moving their nonlinearities inside the definition of $F$.*

## A. CNMA Innovations

CNMA's innovation is connecting the modeling power of neural networks and constraint-solving power of MILP solvers into a learning-from-failure feedback loop in such a way that they do much of the work for us, permitting straightforward, efficient implementations of the following desirable features into a single, cohesive system:

1. **Efficient construction of a surrogate function.** The complexity of neural network is linear in the number of samples.

2. **Efficient constraint-solving without penalty functions.** This feature is enabled by the transformation of neural networks with the ReLU activation function into an equivalent MILP. In addition, constraints are directly expressed in the MILP language and then efficiently solved by industrial-strength MILP solvers such as CPLEX and GUROBI [11, 12].

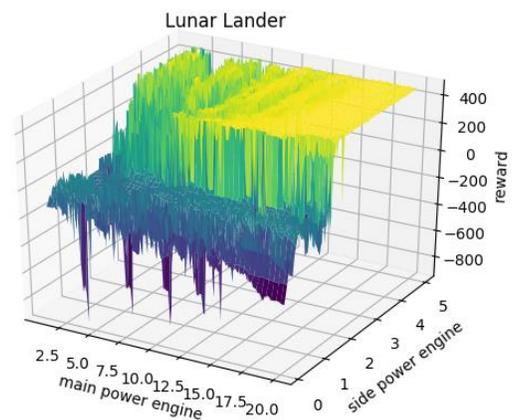

*Figure 2. An example of a function over which CNMA can optimize. It is non-continuous and not even defined at all points. Only two dimensions of the 15 are shown. See Section V.B for details.*

3. **Sample efficiency.** This feature is enabled by CNMA's learning-from-failure feedback loop. A surrogate is learnt as a neural network that is then transformed into an MILP. Using this MILP, a constrained optimization problem is solved and if the solution is unacceptable, it is used to find a new point to sample.

4. **Optimization with discrete and continuous variables.** Function inputs, outputs and constraints can all contain discrete and continuous variables. This feature is enabled by the use of both neural networks and MILP solvers.

5. **Solving constraints whose evaluation itself requires blackbox evaluations.** CNMA handles this by introducing new variables for components of constraints that must be evaluated via blackbox methods and shifts the blackbox estimation to the underlying surrogate that is created. See Section V.H.

6. **Resilience to the failure of blackboxes to compute outputs.** CNMA leverages the ability of neural networks to learn despite missing information. It makes no assumptions about function continuity or smoothness. Figure 2 shows an example of a function CNMA can optimize over.

7. **Parallelism.** A simple parallel version of CNMA, also presented in this paper, improves the efficiency and quality of solutions over the sequential version, and also tries to steer it away from local optima. No restriction is placed on functions that can be optimized in parallel.

CNMA samples points in the domains of the function, and learns a neural network surrogate of the function. By the Universal Approximation Theorem [33], neural networks can approximate any continuous function, although in practice, they are also used to approximate non-continuous, non-smooth functions. CNMA transforms that surrogate into an equivalent MILP [34] and constructs its conjunction with $P(x, y)$. It optimizes $\phi(x, y)$ subject to this conjunction using industrial-strength MILP solvers. It then checks the solution for correctness, i.e., whether it satisfies $P(x, y)$ and whether the objective function is of acceptable value. If so, CNMA outputs the solution. If not, CNMA computes a new training instance from the solution and restarts. This "learning-from-failure" feedback loop has the effect of trying to sample the region of the domain relevant to solving the optimization problem. Thus, it reduces the number of function evaluations by orders of magnitude compared to that needed for learning the function over its entire domain.

A parallel version of CNMA uses multiple agents with each using a different neural network architecture but operating off the same training set. Each independently computes the next best point to sample and adds it to the common training set. The resulting model diversity decreases the chances of getting stuck in local optima. Parallel neural network training, MILP solving and sample evaluation also contribute to improved performance over the sequential version.

Genetic algorithms can be combined with CNMA in a form of hybrid optimization to find solutions that may not be found by one or the other alone. This idea is thoroughly explored in a sister paper of the same title but Part II [28].

CNMA presents a novel method of addressing a major challenge posed in [49]: how to combine inductive and deductive reasoning in the design of cyber-physical systems. In CNMA, inductive reasoning is accomplished by neural networks and deductive reasoning by MILP solvers, with the two tied together in a feedback loop.

The paper is organized as follows. Section II discusses related work, in particular, highlighting the relationship of CNMA with BO/GP. Section III provides the necessary background. Section IV presents sequential and parallel CNMA and illustrates them with a simple example. Section V evaluates CNMA performance for seven nonlinear problems of 8 (two problems), 10, 15, 36 and 60 real-valued dimensions and one with 186 binary dimension. Its performance is compared with that of the `skopt` [35] implementation of BO/GP (abbreviated BO-S) and the scipy.optimize [36] implementation of NM (abbreviated NM-S), and Random Search. BO-S and NM-S are stable, off-the-shelf tools. Note, however, that BO-S did not return a solution for two problems and NM-S did not return one for three. Section VI concludes the paper.

## II. RELATED WORK

Surrogate-based methods are most closely related to CNMA. Of these methods, perhaps the most well-researched is BO/GP [9]. Let the objective function to be maximized be $F(x)$ and let a blackbox be available to evaluate $F(x)$. BO/GP is initialized with a covariance function $k(x, y)$, also called a kernel. BO/GP is also initialized with a set $S$ of samples in the domain of $F$ and their associated values. With this information, BO/GP iteratively updates the posterior distribution of the underlying Gaussian Process, parameterized by $\mu(x)$ and $\sigma(x)$. Intuitively, $\mu(x)$ is the mean of the values at $x$, of all possible functions whose value for any sample $v$ in $S$ is $F(v)$. $\sigma(x)$ is the standard deviation of all these values. These two functions are combined in different ways to create a merit or acquisition function. This function is maximized using an optimization engine such as L-BFGS [37]. The value of $x$ in the solution represents a new point to sample, relevant to the optimization problem that balances exploration with exploitation. It is added to $S$ and the step repeats till a sample satisfying some halting condition is found. One such acquisition function is the Upper Confidence Bound $UCB(x) = \mu(x) + \beta * \sigma(x)$ with $\beta \geq 0$. The construction of $\mu(x)$ and $\sigma(x)$ requires the inversion of the covariance matrix that lists $k(u, v)$ for each pair $(u, v)$ where $u, v$ are samples in $S$. In earlier versions of BO/GP, a matrix inversion method cubic in the number of samples was used [1], although faster methods on GPUs are reported in [15].

If $F(x)$ is to be optimized subject to a constraint $P(x, F(x))$, $P(x, F(x))$ could be modeled as a penalty/barrier function and added to $F(x)$. If $P(x, F(x))$ itself requires a blackbox evaluation, extensions of BO/GP have been proposed [16, 17, 18, 19] that, in their inner loop, find an $x$ for which the likelihood of $F(x)$ being the maximum and that of $P(x, F(x))$ being true is high.

Parallel versions of BO/GP have been proposed in [20, 21, 22]. Some, such as ref. [20], assume function "additivity," i.e., the function can be decomposed into a sum of functions on disjoint subsets of the function domain. Ref. [38] reports the use of a neural network as a surrogate but the surrogate remains inside the Gaussian Processes framework. It is not solved with an MILP solver. Ref. [39] reports a scheme for mixed discrete-continuous variables in BO/GP. To handle failure of function evaluation, ref. [40] reports a scheme for learning problematic areas of the search space and avoiding it.

Like BO/GP, CNMA also builds a surrogate of $F(x)$, say $F_{nn}(x)$, from sampling the blackbox. The selection of the neural network architecture is analogous to selection of the kernel and $F_{nn}(x)$ is analogous to $\mu(x)$. The MILP solver is analogous to an engine such as L-BFGS. It is directly used to find $x$ that maximizes $y$ such that $F_{nn}(x) = y \wedge P(x, y)$. Any solution is used to compute the next point to sample.

Connecting neural networks and MILP solvers in a learning-from-failure feedback loop permits efficient, straightforward implementation of above features: efficient surrogate function construction, sample efficiency, constraint solving without penalty functions, solving blackbox constraints, optimization with discrete and continuous variables, resilience to non-terminating function evaluations, and parallelism.

CNMA does not compute the variance of the surrogate it learns. Effectively, its merit function is $UCB(x)$ with $\beta = 0$.

There is, thus, a risk that it could get stuck in local optima finding more and more points around the current optima. CNMA provides two methods for trying to avoid this problem and searching globally. The first is to use a constraint stating that the objective function is above a threshold. Then, the MILP solver will not produce solutions for which the objective is below the threshold. The second is to introduce model diversity to reduce the chances of different models computing the same local optima. Model diversity is a byproduct of parallel CNMA whose multiple agents create their own models. Independently, parallel CNMA also contributes to improved performance via parallel neural network training, MILP solving and sample evaluation.

If the current surrogate is not good enough then $F_{nn}(x)$ subject to $P(x)$ may have no solution. In that case, CNMA generates a random point and restarts. How to mitigate such randomness is one problem of current research. Other future research problems include introducing additional diversity, e.g., by use of bootstrapping, multi-function CNMA, finding an appropriate initial neural network architecture and adapting that architecture as new samples are created, e.g., via the use of network compression [41].

We now discuss CNMA in detail.

## III. BACKGROUND

A mixed-integer linear constraint is of the form $a_0 * x_0 + \ldots + a_k * x_k \leq b$ where $a_i, b$ are real-valued constants and the $x_i$ are real-valued or integer-valued. An MILP is a set of such constraints with a linear objective function $\phi(v_0, \ldots, v_m)$ where each $v_i$ is a variable appearing in a constraint. An MILP solver finds values of all variables in the program optimizing the function while satisfying all constraints. It makes no distinction between input and output variables.

A neural network is a set of layers with each layer consisting of a set of neurons. In the fully-connected neural network used in CNMA, each neuron in a layer is connected to each neuron in the previous layer. Such a network has one input layer, one output layer and zero or more hidden layers. When the values of neurons in the input layer are initialized, they are propagated forward to compute values of all neurons. The output layer can have multiple neurons allowing modeling of multi-output functions. Associated with the edge between two neurons is a weight. Associated with each neuron is a bias or intercept. The value of a hidden-layer neuron is a linear combination of its bias, values in the previous layer and weights of connecting edges, but passed through an activation function. Activation functions give neural networks the power to model nonlinear functions. We use the ReLU activation function $max(x, 0)$ because it can be converted into an MILP constraint using the big-M method [34]. By also modeling the overall system requirement as another mixed integer linear constraint, scalable MILP solvers can be used to efficiently solve the neural network along with the requirement. To allow neural networks to model negative outputs, no activation function is applied at the output layer.

We now illustrate the above plan with a short example. To model the equation $y = max(x, 0)$ as an MILP, select a large number $M$ and let an integer $d \in \{0, 1\}$. Then, $y = max(x, 0)$ is equivalent to $(y \geq 0 \; \wedge y \geq x \wedge y \leq x + M * d \wedge y \leq$

$M(1 - d))$.

***Proof of correctness of tranformation of ReLU into MILP.***
To see how the MILP $(y \geq 0 \; \wedge y \geq x \wedge y \leq x + M * d \wedge y \leq M(1 - d))$ is equivalent to $y = max(x, 0)$, consider two cases. In the first case, let $d = 0$. The MILP simplifies to $(y \geq 0 \; \wedge y \geq x \wedge y \leq x \wedge y \leq M)$. Because $M$ is a large number, $y \leq M$ is satisfied trivially and the MILP further simplifies to $(y \geq 0 \; \wedge y = x)$ which can also be written as $(y = x \wedge x \geq 0)$. In the second case, let $d = 1$. The MILP simplifies to $(y \geq 0 \; \wedge y \geq x \wedge y \leq x + M \wedge y \leq 0)$. Because $M$ is a large number $y \leq x + M$ is satisfied trivially and the MILP further simplifies to $(y = 0 \; \wedge y \geq x)$ which can also be written as $(y = 0 \; \wedge x \leq 0)$. By combining these two cases, the MILP is equivalent to $y = x$ if $x \geq 0$, otherwise $y = 0$. Finally, this is equivalent to $y = max(x, 0)$.

We now show how to model a whole neural network as an MILP.

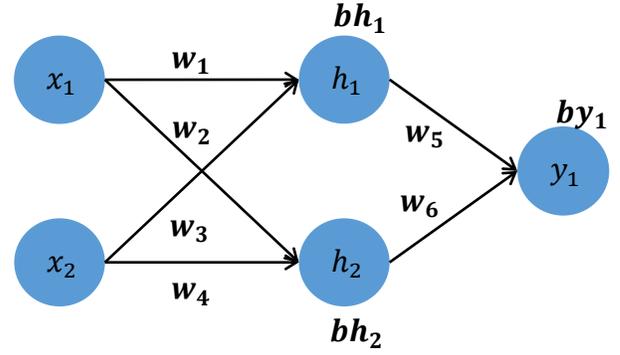

*Figure 3. A fully-connected neural network.*

The neural network in Figure 3 uses the ReLU activation function. Each blue circle represents a single neuron. $x_1$ and $x_2$ represent the two inputs that get fed into the neural network. Neurons $h_1$ and $h_2$ represent the values of each neuron in the hidden layer after the activation function is applied, and $y_1$ represents the output of the neural network. $w_1$ to $w_4$ represent the weights of the first layer and $w_5$ and $w_6$ represent the weights of the output layer. Finally, $bh_1$, $bh_2$, and $by_1$ represent the bias term of the respective neuron. The value of each neuron can then be computed by the following equations:

$$h_1 = max(w_1 * x_1 + w_3 * x_2 + bh_1, 0) \wedge$$
$$h_2 = max(w_2 * x_1 + w_4 * x_2 + bh_2, 0) \wedge$$
$$y_1 = w_5 * h_1 + w_6 * h_2 + b_{y1}$$

Letting $M = 100000$ and $d_1, d_2$ integers $\in \{0, 1\}$, the equivalent MILP for this network, $nn\_milp$, is:

$$nn\_milp = (h_1 \geq 0 \; \wedge$$
$$h_1 \geq w_1 * x_1 + w_3 * x_2 + bh_1 \wedge$$
$$h_1 \leq w_1 * x_1 + w_3 * x_2 + bh_1 + 100000 * d_1 \wedge$$
$$h_1 \leq 100000 * (1 - d_1) \wedge$$

$$h_2 \geq 0 \; \wedge$$
$$h_2 \geq w_2 * x_1 + w_4 * x_2 + bh_2 \wedge$$
$$h_2 \leq w_2 * x_1 + w_4 * x_2 + bh_2 + 100000 * d_2 \wedge$$
$$h_2 \leq 100000 * (1 - d_2) \wedge$$

$$y_1 = w_5 * h_1 + w_6 * h_2 + by_1)$$

Let the above neural network model a blackbox function $F(x_1, x_2) = y_1$ and let $\phi(x_1, x_2, y_1), P(x_1, x_2, y_1)$ be an MILP objective function and constraint, respectively. Then, to optimize $\phi(x_1, x_2, y_1)$ subject to $F(x_1, x_2) = y_1 \wedge P(x_1, x_2, y_1)$, we can use an MILP solver to optimize $\phi(x_1, x_2, y_1)$ subject to $nn\_milp \wedge P(x_1, x_2, y_1)$.

A central innovation of CNMA is that the neural network does not have to model $F$ exactly to solve the optimization problem. To model it exactly would require an astronomical number of samples from $F$'s domain. Instead, CNMA tries to choose samples that are relevant to solving the optimization problem and are thus a tiny fraction of the number required for accurate modeling. After each sample, the neural network surrogate is reconstructed and the optimization problem solved again. The solution is used to compute the next promising sample and the step repeated till an acceptable solution is found.

## IV. METHODOLOGY

CNMA solves the problem of finding $x, y$ that optimize $\phi(x, y)$ such that $F(x) = y \wedge P(x, y)$ where $F$ is a nonlinear function, $x, y$ are vectors of discrete and continuous variables, $\phi$ is a linear function and $P$ is a linear constraint. $F$ is called the forward function. As shown in Figure 4, CNMA uses a Sample Generator to sample points in $F$'s domain, evaluates those points and creates a training set. These points are inputs to the neural network (NN) ReLU Regression Engine that outputs a neural network $nn$. This is transformed into an equivalent MILP $milp$ by the NN-MILP transformer. $milp$ is a surrogate or model of $F$ based on current $samples$. An MILP solver solves $\phi(x, y)$ such that $milp \wedge P(x, y)$ is true. If a solution is not found, the Sampling Engine is called upon to extend $samples$ with a new one in the hope of improving upon the current surrogate, and CNMA restarts. Otherwise, let $(x^*, \hat{y})$ be a solution. It is then checked for correctness, i.e., whether $P(x^*, F(x^*))$ holds. If it does, then the value of the objective function $\phi(x^*, F(x^*))$ is checked for acceptability, e.g., whether it is above or below a desired threshold or whether the evaluation budget has been reached. If the solution is acceptable, $(x^*, F(x^*))$ is output and CNMA halts. Otherwise, $(x^*, F(x^*))$ is added to $samples$. If $P(x^*, F(x^*))$ is false, then $(x^*, F(x^*))$ is also added to $samples$. Now, CNMA restarts. Note that CNMA can be used in pure constraint satisfaction mode by letting $\phi(x, y) = 0$ and in pure optimization mode by letting $P(x, y) = true$.

The addition of $(x^*, F(x^*))$ if $P(x^*, F(x^*))$ does not hold is a form of learning from the failure to produce a surrogate of $F$ that intersects $P$. It has the effect of trying to restrict the sampling to only the part of $F$'s domain that is relevant to the satisfaction of $P$. Thus, the sampling of $F$ is reduced by many orders of magnitude over the fine-grained sampling needed to learn $F$ over its entire domain. Even if $P(x^*, F(x^*))$ does hold, adding it to $samples$ in the hopes of improving upon $\phi(x)$ also restricts the sampling.

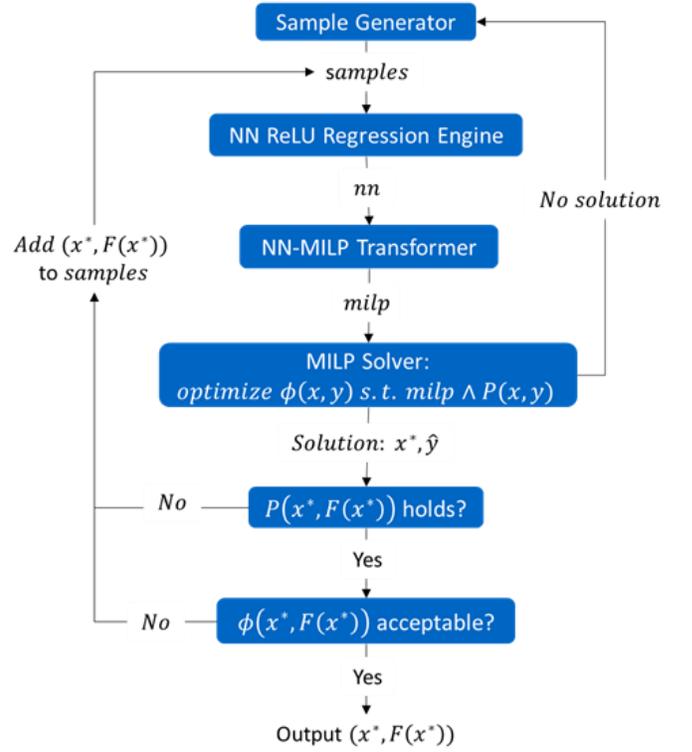

*Figure 4. Sequential CNMA overview*

For clarity, the flowchart in Figure 4 (and Algorithms 1, 2 and Figure 5) omit an important case: the failure to evaluate $F$ for a given input either in the creation of the initial samples or in the evaluation of $P(x^*, F(x^*))$. In this case, CNMA just calls upon the Sample Generator to generate a new sample and restarts. Neural networks are resilient to missing data.

Algorithm 1 precisely defines the above plan.

---

**Algorithm 1** CNMA, Single Forward Function

**Input:** a problem definition of the form $\max_{x,y} \phi(x, y)$ s.t. $y = F(x) \wedge P(x, y)$, a method to randomly sample candidate solutions $x \in X$, maximum number $N$ of CNMA iterations. $\phi, P$ are linear, $x, y$ are vectors of discrete and continuous variables, and $F$ is a potentially nonlinear function available as a blackbox.

**Output:** a solution, $(x^*, y^*)$, to the above problem

**function** CNMA($\phi, F, P, X, N$)
  $samples \leftarrow \left\{\left(x_i, F(x_i)\right)\right\}_{i=1,2,\cdots,n}$ where $x_i$ denotes a random sample drawn from $X$
  $solutions \leftarrow$ empty list
  **for** $i = 1, 2, \cdots, N$:
    $nn \leftarrow$ a fully-connected ReLU regression network, i.e., with identity activation for the last layer, which takes in as input a vector $x \in X$ and attempts to predict $f(x)$; use $samples$ to train this neural network
    $milp \leftarrow$ the mixed-integer linear program: $\max_{x,y} \phi(x, y)$ s.t. $nn\_to\_milp(nn) \wedge P(x, y)$
    $x^*, \hat{y} \leftarrow$ potentially infeasible solution to $milp$, obtained via an MILP solver
    **if** $(x^*, \hat{y})$ is a feasible solution to $milp$:

$y^* \leftarrow F(x^*)$ // if $F$ does not terminate within some time limit then $F$ returns $\infty$

> **if** $y^*$ is finite:
>> **if** $P(x^*, y^*)$ is satisfied:
>>> Append $(x^*, y^*)$ to *samples*
>>> Append $(x^*, y^*)$ to *solutions*
>> **else:**
>>> Append $(x^*, y^*)$ to *samples*
>> **endif**
> **endif**
> **else:** // the solution to *milp* is infeasible
>> Append randomly drawn sample(s) $\{x_i, F(x_i)\}_{i=1,2,\cdots}$ to *samples*
> **endif**
> **return** the best solution from *solutions*, sorting by $\phi(x, y)$

**endfunction**

It is possible that CNMA could get stuck in local optima. There are two methods of steering it away from them. The first is adding a constraint on an upper or lower bound of the objective function so that the MILP solver finds solutions satisfying that constraint. The second is to use multiple concurrent copies of CNMA, each producing a different surrogate and therefore adding different "good" points to *samples*. By pooling together these good points, the chances of finding the true optima can be improved. This method is a byproduct of Parallel CNMA that parallelizes sampling, training and solving to improve the quality and performance of the sequential version.

## A. Parallel CNMA

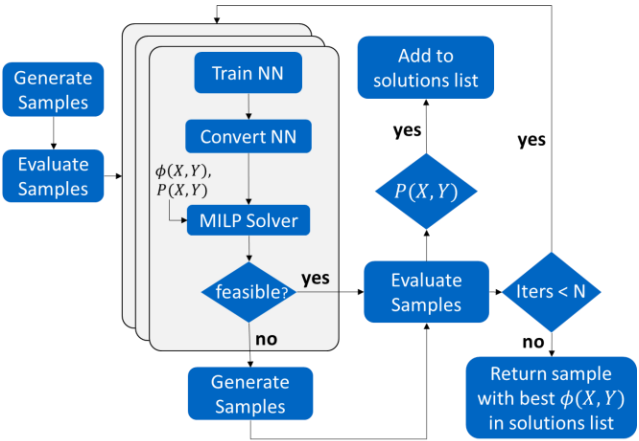

*Figure 5. Parallel CNMA overview*

As shown in Figure 5, the CNMA algorithm can also be parallelized by having multiple instances of CNMA run simultaneously while periodically sharing information with each other. During each iteration, each CNMA solver uses the existing evaluated samples as training data to create a surrogate model of the forward function, $F$. Each solver may use a different neural network architecture or initialize the neural network weights differently, causing each solver to end up with a unique MILP problem to solve and produce a different point to sample next. Because there are many different neural networks that all may fit the training data, running CNMA in parallel allows us to explore more of these diverse surrogate models in a single iteration. At the end of each iteration, all evaluated samples are shared among the solvers. See Algorithm 2 for a precise definition of Parallel CNMA.

---

**Algorithm 2** Parallel CNMA, Single Forward Function

**Input:** a problem definition of the form $\max\limits_{x,y} \phi(x,y)$ s.t. $y = F(x) \land P(x,y)$, a method to randomly sample candidate solutions $x \in X$, maximum number $N$ of CNMA iterations, $M$ parallel solvers. $\phi, P$ are linear, $x, y$ are vectors of discrete and continuous variables, and $F$ is a potentially nonlinear function available as a blackbox.

**Output:** a solution, $(x^*, y^*)$, to the above problem

**function** CNMA$(\phi, F, P, X)$

> $samples \leftarrow \left\{\left(x_i, F(x_i)\right)\right\}_{i=1,2,\cdots,n}$, where $x_i$ denotes a random sample drawn from $X$
> $solutions \leftarrow$ empty list
> **for** $i = 1, 2, \cdots, N$
>> $candidate\_solutions \leftarrow$ empty list
>> **parallel for** $j = 1, 2, \cdots, M$
>>> $nn \leftarrow$ a fully-connected ReLU regression network (with any arbitrary architecture), which takes in as input a vector $x \in X$ and attempts to predict $F(x)$; use *samples* to train this neural network.
>>> $milp \leftarrow$ the mixed-integer linear program: $\max\limits_{x,y} \phi(x,y)$ s.t. $nn\_to\_milp(nn) \land P(x,y)$
>>> $x^*, \hat{y} \leftarrow$ solution to *milp*, obtained, e.g., via an MILP solver
>>> **if** $(x^*, \hat{y})$ is feasible:
>>>> $candidate\_solutions[j] \leftarrow (x^*, \hat{y})$
>> $evals \leftarrow$ in parallel, compute $\left(x_i, F(x_i)\right)$ for each $x_i \in candidate\_solutions$
>> $n_{invalid} \leftarrow 0$
>> **for** $j = 1, 2, \cdots,$ size_of($evals$)
>>> $x^*, \hat{y} \leftarrow candidate\_solutions[j]$
>>> **if** $(x^*, \hat{y})$ is infeasible:
>>>> $n_{invalid} \leftarrow n_{invalid} + 1$
>>> **else**:
>>>> Append $(x^*, \hat{y})$ to *samples*
>>>> **if** $P(x^*, \hat{y})$:
>>>>> Append $(x^*, \hat{y})$ to *solutions*
>>> In parallel, append $n_{invalid}$ randomly drawn sample(s) $\left(x_i, F(x_i)\right)$ to *samples*
> **return** the best solution from *solutions*, sorting by $\phi(x, y)$

**endfunction**

---

## B. Illustrating CNMA

The Rastrigin function is a common benchmark problem for optimization methods because it is highly nonlinear and has many local optima. We illustrate three different ways in which CNMA can be used to solve the optimization problem:

$$\max\limits_{x \in [-5.12, 5.12]} F(x) = 10 + x^2 - 10 * \cos(2 * \pi * x)$$

The true maximum of F(x) is 40.353 at $x = -4.522$ and $x = 4.522$.

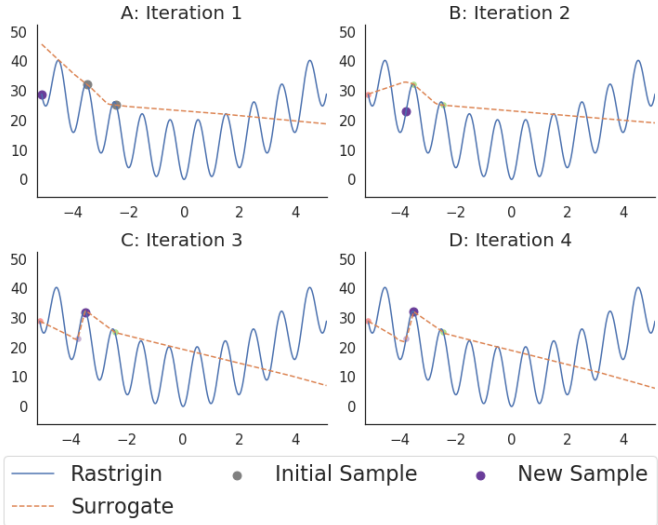

*Figure 6. Maximizing Rastrigin. A: During the first iteration the surrogate function predicts that the maximum is at -5.12. We evaluate Rastrigin at this point and the new sample, shown as a purple dot, is added to the training set. B: During iteration 2, the surrogate incorrectly predicts where the maximum value of the function is. The new sample, shown as a purple dot, is added to the training set. C: During iteration 3, the surrogate incorrectly predicts where the maximum value of the function is. The new sample, shown as a purple dot, is added to the training set. D: During iteration 4, the surrogate incorrectly predicts where the maximum value of the function is. The new sample, shown as a purple dot, is added to the training set.*

Figure 6 shows the progression of CNMA at each iteration solving: maximize $\phi(x, y) = y$ s.t. $F(x) = y \land -5.12 \leq x \leq 5.12$. CNMA first generates two initial samples in the domain of $x$ to create the training set:

| $x$ | $F(x)$ |
|---|---|
| $-3.495$ | 32.210 |
| $-2.436$ | 25.161 |

CNMA then learns a neural network from this set to create a surrogate of $F(x)$. Figure 6 shows the surrogate function plotted in orange. CNMA converts this neural network into an MILP and uses an MILP solver to solve it along with $P(x, y) = -5.12 \leq x \leq 5.12$ such that $y$ is maximized.

During the first iteration, the solution found is $x = -5.12$. This is where the maximum of the orange curve is within $x$'s domain. After checking the solution against the correct definition of $F(x)$, the dot shown in purple is added to the training set and a new neural network is trained during Iteration 2. While CNMA is able to find a local maximum after just three iterations and five function calls, CNMA gets stuck here and a global maximum is not found even after 100 iterations.

To help CNMA explore outside this local maximum, we can use Parallel CNMA to add neural network diversity. Figure 7 shows how Parallel CNMA solves the same problem but is able to find the global maximum by using 10 different CNMA workers simultaneously. During the first iteration, 10 neural networks, each initialized with different weights, are trained on the same two initial samples. This creates 10 different MILP problems to be solved. Because each neural network is slightly different, each MILP problem produces a different solution. The eighth neural network predicts that the maximum value of the function is at $x = -4.571$, which is close to one of the true maxima at $x = -4.5229$. Since all samples are shared among all parallel workers, by the next iteration each neural network predicts that the maximum is very close to the true maximum at $x = -4.522$. After only three iterations, the best solution found is $x = -4.521$, which has an objective function value of $y = 40.353$.

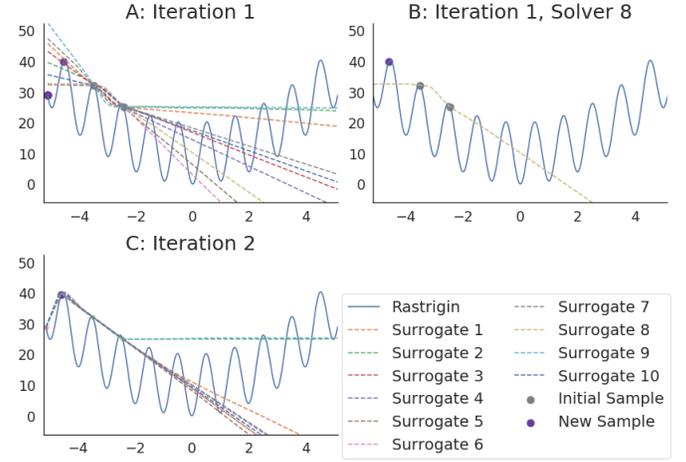

*Figure 7. Maximizing Rastrigin with multiple CNMA solvers. A: During Iteration 1, 10 surrogates are trained. B: The eighth surrogate trained in Iteration 1 has a maximum at $x=-4.571$. This is very close to the true maximum of the Rastrigin function. C: During Iteration 2, all 10 surrogates accurately predict the maximum value of the Rastrigin function.*

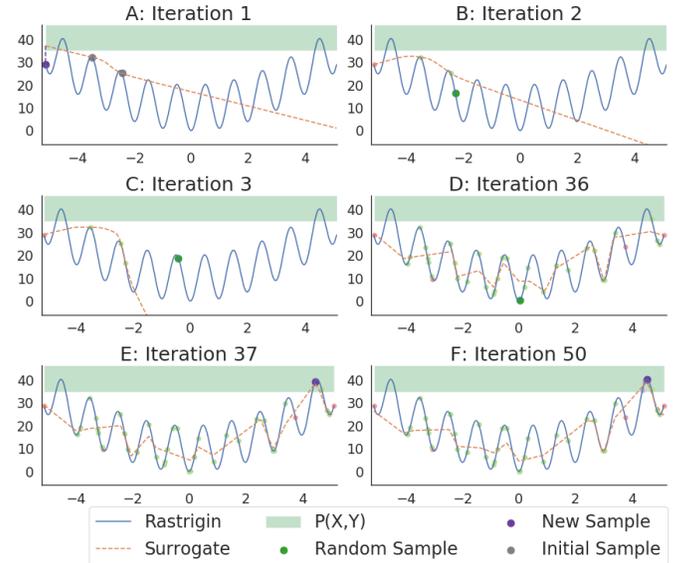

*Figure 8. Maximizing Rastrigin with constraints. A: In Iteration 1 the surrogate overlaps with the constraint, $P(X,Y)$. The point shown in purple is added to the training set. B: In Iteration 2 the surrogate does not overlap with $P(X,Y)$. A random sample, shown in green, is added to the training set. C: In Iteration 3 the surrogate does not overlap with $P(X,Y)$. A random sample, shown in green, is added to the training set. D: Again, in Iteration 36, the surrogate does not overlap with $P(X,Y)$. A random sample, shown in green, is added to the training set. E: In Iteration 37 the surrogate does overlap with $P(X,Y)$. The dot shown in purple is added to the training set. F: In Iteration 50, the maximum value is found at $x=4.523$.*

Another way to try to ensure CNMA does not get stuck in local maxima is to add a constraint to ensure that the objective function value is above a desired threshold. In Figure 8, we depict the progression of CNMA with the added constraint $y \geq 35$. During Iteration 1, the MILP solver finds $x = -5.12$ because it maximizes the surrogate in orange and $P(x, y)$ (shaded in light green) is satisfied. However, when the solution is evaluated by the true function (shown in purple), $P(x, y)$ no longer holds. In the second iteration, there is no $x$ such that the surrogate and $P(x, y)$ overlap. Thus, the MILP problem is infeasible and a random sample is evaluated instead (depicted as a green dot). Many random samples continue to be generated until Iteration 37, where the surrogate function again overlaps with $P(x, y)$. After evaluating the MILP solution through the true function, this time $P(x, y)$ is satisfied. This causes the surrogate function to continue to overlap with $P(x, y)$ during subsequent iterations. The solution $x = 4.523$ with objective function $y = 40.353$ is found after 50 iterations and 52 function calls (2 initial + one each for 50 iterations).

## V. Experimental Results

For seven nonlinear design problems of 8 (two problems), 10, 15, 36, 60 real-valued dimensions and one with 186 binary dimension, we compare the performance of CNMA with the `skopt [35]` implementation of BO/GP (abbreviated BO-S), the `scipy.optimize [36]` implementation of NM (abbreviated NM-S), and with Random Search. We evaluated CNMA with 1, 5, and 10 solvers in parallel. For neural networks, we use the `scikit-learn package [42]`. We use a commercial MILP solver. All of these packages are stable-off-the-shelf.

For each problem, we allocate a fixed time budget that is, in our estimate, the longest a user would wait for a solution to that problem.

For each problem, we compare the optimization engines against two metrics. The first metric is the best value of the objective function computed within the problem's time budget. It is evaluated by plotting the improvement of the objective function value against time, and comparing the best values from each engine.

The second metric is the minimum number of function evaluations needed to produce the best value of the objective function within the problem's time budget. This metric is intended to capture the idea of "sample efficiency" since it can be quite expensive to evaluate blackboxes. This metric is evaluated by plotting the improvement of the objective function against the number of function evaluations, and comparing the minimum number of evaluations needed by each engine to produce the best value of the objective function.

We also compare Random Search by randomly generating the maximum number of samples used by any one of the optimization methods which found solutions.

For BO-S, we use the expected improvement (EI) acquisition function, and the Matérn kernel. The following hyperparameters are automatically tuned by BO-S: (1) all kernel length scales, (2) covariance amplitudes, and (3) parameters of the i.i.d. Gaussian noise added to the kernel.

For CNMA with 1 solver, we use a neural network architecture with two hidden layers of 35 and 10 neurons. For CNMA with 5 solvers, we use the same neural network architecture in addition to ones with single hidden layers of 10, 30, 35, and 50 neurons. We use the same architectures (each repeated twice) for CNMA with 10 solvers.

For BO-S and NM-S, we model constraints as a penalty function.

Note that BO-S did not return a solution for two problems and NM-S did not return one for three. These problems are high-dimensional, have both discrete and continuous variables, and their blackboxes do not return outputs for some inputs. For other problems, CNMA improves the performance over BO-S, NM-S and Random Search by 1%-87%.

Some comparative visualizations are available at https://collab.perspectalabs.com/nonlinearbenchmarks/.

### A. Designing a Wave-Energy-Propelled Boat

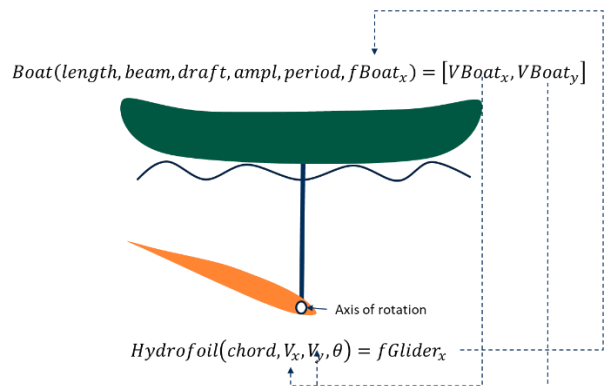

$Boat(length, beam, draft, ampl, period, fBoat_x) = [VBoat_x, VBoat_y]$

$Hydrofoil(chord, V_x, V_y, \theta) = fGlider_x$

*Figure 9. Structure of wave energy propelled boat. Note the recursive relationship between the input and output variables of Boat and Hydrofoil*

This example shows the ability of CNMA to optimize with functions that may fail to return a value for some inputs, as can happen with the `xfoil [29]` computational fluid dynamics simulator used here. Figure 9 indicates how the rise and fall of a boat floating on a wave pulls and pushes at a hydrofoil below, causing a rotation about an axis. This rotation generates forward force during both the upward and downward wave motion; much like when swimmers flap flippers in a pool, their body is propelled forward. The marine robot is inspired by the Wave Glider [43]. The design problem is to compute the dimensions of the boat and hydrofoil which will maximize the steady-state forward sailing speed for a given wave condition. The equilibrium constraints are that the force generated by the hydrofoil equals that applied to the boat and that the glider and boat velocities are equal. An additional constraint is that the magnitude of the horizontal velocity be higher than that of the vertical velocity. Note the recursive relationship between the variables: force is an output of $Hydrofoil$ but an input to $Boat$ whereas velocities are outputs of the latter and inputs to the former.

The boat is modeled with two functions. The first is $Hydrofoil(ch, V_x, V_y, \theta) = fGlider_x$ that computes the forward force output by a hydrofoil of length $ch$, moving through water at velocity $V_x, V_y$ at an angle of attack $\theta$. This is the force it applies to the boat. It is implemented with the computational fluid dynamics package `xfoil`. The second function is ($length, beam, draft, ampl, period, fBoat_x$) = $[VBoat_x, VBoat_y]$. It outputs the steady-state forward speed of a boat given its 3 dimensions: $length, beam, draft$, the amplitude $ampl$ and $period$ of the wave, and the forward force $fBoat_x$ applied to it by the hydrofoil. This function is computed by a program encoding a solution to a differential equation. The first two equilibrium constraints are enforced by eliminating $V_x$ and $V_y$ and consolidating the two functions into the CNMA forward function $F(x) = y$ where:

$$x = [length, beam, draft, ampl, period, fBoat_x, ch, \theta]$$
$$y = [VBoat_x, VBoat_y, fGlider_x]$$

$F$ calls $Boat$ to compute $VBoat_x$ and $VBoat_y$ and then inputs them to $Hydrofoil$ to compute $fGlider_x$. The third equilibrium constraint is now $fBoat_x = fGlider_x$. To tolerate small force differences, the equality is modeled as $|fBoat_x - fGlider_x| \le \epsilon * fBoat_x$ where $\epsilon$ is set to 5%. Note that a constraint with an absolute value can be modeled as a pair of linear constraints [44]. Finally, $P(x, y) = (|fBoat_x - fGlider_x| \le \epsilon * fBoat_x \wedge VBoat_x \ge VBoat_y)$ and $\phi(X, Y) = VBoat_x$. In the optimization problem, $ampl$ and $period$ are set to 4 and 7, respectively.

With 30 initial samples and a budget of 2500 seconds, with 1, 5, and 10 solvers, CNMA is able to find a solution with a speed of 3.9036, 3.8993, and 3.8999 m/s, respectively. BO-S and NM-S are unable to find any feasible solutions given the same time budget. Out of 3890 randomly samples, none of them are valid solutions. Figure 10 and Figure 11 show the performance of CNMA with respect to time and number of function evaluations.

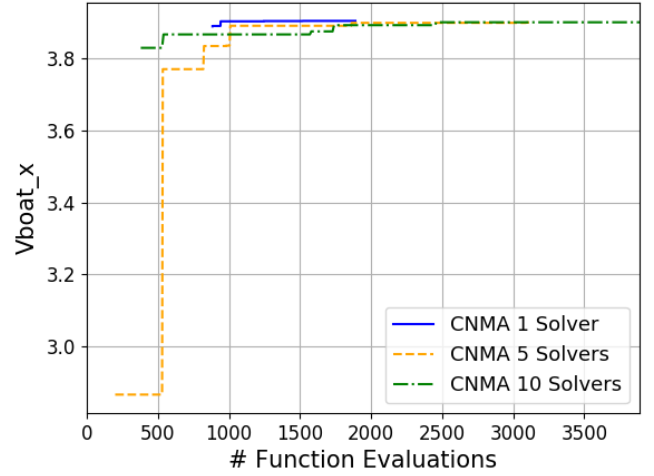

*Figure 11. Boat velocity in the x direction vs. number of function evaluations for Wave-Energy-Propelled Boat. BO-S, NM-S, and Random Search are not shown since they are not able to find valid solutions. After 1750 function evaluations all three CNMA runs find solutions with a speed above 3.89 m/s.*

### B. OpenAI Gym's LunarLander: Robot System-Controller Codesign

Robots and their controllers are often designed separately. This can lead to design inconsistency whose resolution can be time-consuming. If robots can be efficiently reconfigured and their controllers efficiently designed, we raise the possibility of system-controller codesign by solving a single optimization problem that encodes system and controller constraints and objectives. The lander we use, as depicted in Figure 12, is part of the robotic benchmarks at OpenAI/Gym [45]. We assume that the controller is proportional-integral-derivative (PID)-based. Hence, the controller design problem reduces to finding optimal values for the PID coefficients. From a mothership, the module is ejected with a certain force and then its engines fire both vertically and horizontally to guide it towards landing on the flat pad between two flagpoles. Our goal is to compute the system and controller design and initial position and force that would maximize the reward while satisfying constraints on a successful landing, time to land and fuel usage.

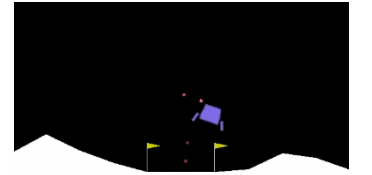

*Figure 12. Lunar lander attempting to land on flat terrain between flagpoles*

The input vector to the CNMA forward function $F$ is $x = [mep, \; sep, \; la, \; ld, \; lw, \; lh, \; lst, \; kp\_alt, \; kd\_alt, \; kp\_ang, \; kd\_ang, \; initial\_x, initial\_y, initial\_fx, initial\_fy]$. The variables $mep, sep, \; la, \; ld, \; lw, lh \; lst$ are system design parameters denoting, respectively, main engine power, side engine power, leg away length, leg down length, leg width, leg height, and leg spring torque. The variables $kp\_alt, kd\_alt, \; kp\_ang, \; kd\_ang$ are controller design parameters denoting, respectively, the P and D values for the vertical and horizontal engine controllers. The I coefficient is set to 0. The variables $initial\_x, initial\_y,$ $initial\_fx, initial\_fy$ are initial condition parameters, denoting, respectively, the initial position and force at the time of lunar lander ejection from the mother ship. The forward function $F(x) = y$ simulates the trajectory of the lander defined by its system design parameters, PID parameters and the initial conditions. It runs for a fixed number of time steps or until the lander lands or crashes. At each time step, it measures the "error," i.e., the distance between its current position and the landing point, and using the PID values computes engine firing actions to guide the lander towards the landing point. It outputs $y = [fuel, time, success, reward]$ denoting, respectively, the fuel used, time taken to land, whether the landing is safe, and the reward, a measure of the quality of the landing, as defined by OpenAI/Gym. The codesign problem is now to maximize the objective function $\phi(x,y) = reward$ subject to $F(x) = Y \land P(x,y)$ where $P(x,y) = (success = 1 \land fuel \leq 75 \land time \leq 10)$.

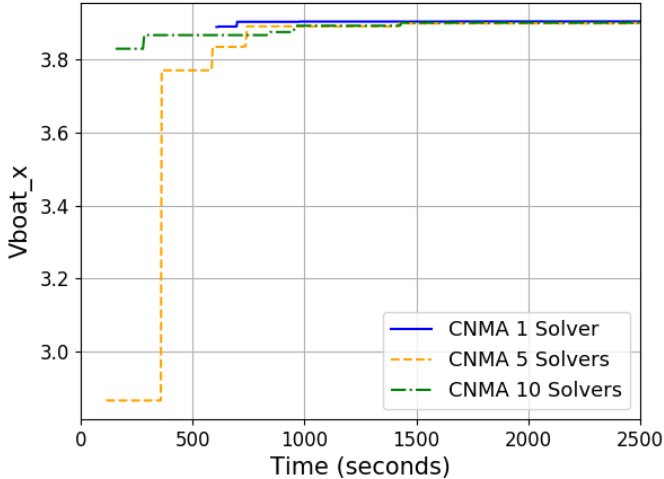

*Figure 10. Boat velocity in the x direction vs. time for Wave-Energy-Propelled Boat. BO-S and NM-S are not shown since these are not able to find valid solutions. After 1000 seconds all three CNMA runs find a speed with a speed above 3.89 m/s.*

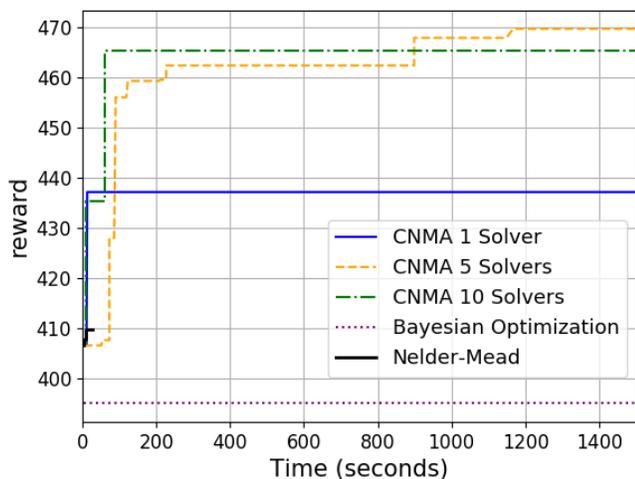

*Figure 13. Reward vs. time for Lunar Lander. After 100 seconds all three CNMA runs find better solutions than BO-S and NM-S. Note that NM-S stops after 28 seconds due to early convergence.*

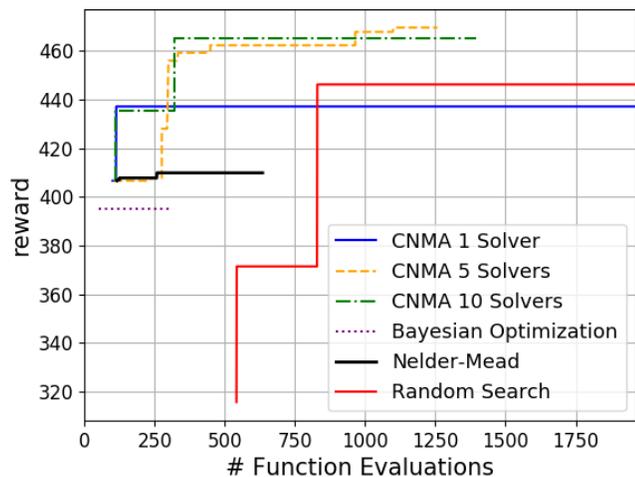

*Figure 14. Reward vs. number of function evaluations for Lunar Lander. When given the same number of function evaluations, CNMA either matches or outperforms both BO-S and NM-S. CNMA with 5, and 10 solvers always outperforms Random Search with a function evaluation budget of less than 1750.*

With 100 initial samples and a time budget of 1500 seconds, CNMA finds solutions with rewards of 437.139, 469.589, and 465.136 with 1, 5, and 10 solvers, respectively. In the same amount of time, BO-S only finds a solution with a reward of 395.118 and NM-S only finds a solution with a reward of 409.857. Out of 1966 random samples, the best solution which meets the constraints has a reward of 446.199. Figure 13 and Figure 14 show the performance of CNMA, BO-S, NM-S and Random Search with respect to time and number of function evaluations.

## C. Optimizing Hexapod Gaits

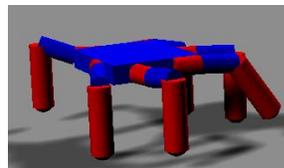

*Figure 15. A hexapod*

We now address a problem inspired by the work of [46] for adapting robot gait to failures in the field. The robot is six-legged with each leg consisting of three segments. Associated with each leg $i$ is a vector of six parameters $(\alpha i_1, \alpha i_2, \varphi i_1, \varphi i_2, \tau i_1, \tau i_2)$ with each $\alpha, \phi, \tau \in [0,1]$. These parameters determine, respectively, the amplitude phase and duty cycle of the walking signal sent to the first two legs every 30 ms. The walking signal for the third segment is the inverse of that for the second so does not need independent control parameters. The hexapod (as shown in Figure 15) controller is defined by the six parameters for each leg for a total of 36 parameters, and fully determines the hexapod gait. Using the hexapod simulator in [46], we define a CNMA forward function $hexapod(x) = y$ that takes a controller $x = [c_0, .., c_{35}]$ as input, simulates the hexapod gait for 5 seconds and outputs a vector $y = [speed_x, b_1, b_2, b_3, b_4, b_5, b_6]$ where $speed_x$ is the hexapod's $x$-axis displacement in meters divided by 5.0 and each $b_i$ is the fraction of the time leg $i$ is in contact with the ground. Hexapod speeds above 0.20 m/sec are hard to find with Random Search [46].

If a hexapod leg is broken then we would like to find a new controller that can achieve a speed of above 0.20 m/sec while satisfying any new constraints on the movement. Let us assume leg 1 is broken. Then, we might constrain its contact with the ground to be the least of that of all the legs. The problem is now: maximize $speed$ subject to $hexapod([c_0, .., c_{35}]) = [speed, b_1, .., b_6] \land b_1 \leq b_2 \land$ $b_1 \leq b_3 \land b_1 \leq b_4 \land b_1 \leq b_5 \land b_1 \leq b_6$ and stop when the speed is close enough to 0.20 m/sec. As can be seen above, the baseline controller does not satisfy the constraint.

We can try searching over $[0,1]^{36}$ for a new controller. However, another option is to search just in the neighborhood of the baseline controller. We change the bounds for each field in the baseline controller to be within 0.1 of its current value, subject to the lower bound being at least 0 and the upper bound at most 1.

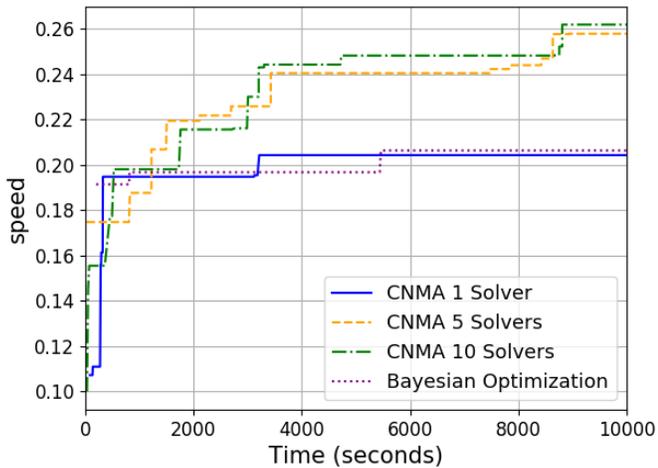

*Figure 16. Speed vs. time for Hexapod. NM-S stops after 207 seconds due to early convergence. NM-S is not shown because it is not able to find any valid solutions. After 2000 seconds CNMA with 5 and 10 solvers outperforms BO-S.*

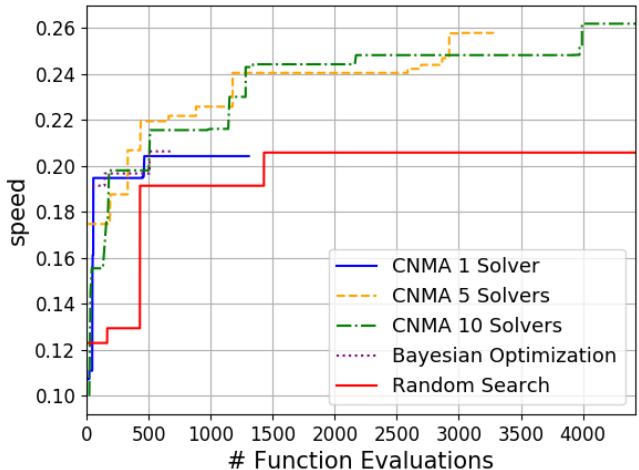

*Figure 17. Speed vs. number of function evaluations for Hexapod. NM-S is not shown because it is not able to find valid solutions. CNMA outperforms Random Search for any given budget of function evaluations less than 4422. CNMA with 5 and 10 solvers matches or outperforms BO-S given a budget of function evaluations between 400 and 600.*

With two initial samples and a time budget of 10000 seconds, CNMA finds solutions with speeds of 0.2043, 0.2579, and 0.2619 m/sec with 1, 5, and 10 solvers, respectively. BO-S finds a solution with a speed of 0.2064 m/sec while NM-S is not able to find any valid solutions. Out of 4422 randomly generated samples, the best solution has a speed of 0.2058 m/sec. Figure 16 and Figure 17 show the performance of CNMA, BO-S, NM-S, and Random Search with respect to time and number of function evaluations.

### D. Acrobot Design

The Acrobot [45], as shown in Figure 18, is a two-link robot arm with a single actuator placed at the elbow. Initially, the links hang downwards. The Acrobot's goal is to execute a series of actions that vertically orients and balances both links. The Acrobot problem is well-studied, and is known to be challenging to solve. In Figure 18, system design variables $m_i$, $l_i$, and $l_{c_i}$ denote, respectively, the mass, moment of inertia, and center of mass location of link $i$. The CNMA function $F$ takes as input a system design vector $x = [m_1, m_2, I_1, I_2, l_{c_1}, l_{c_2}, l_1, l_2]$, runs iterative linear-quadratic regulator (LQR) as our controller on an Acrobot system with this vector and returns $t_{stabilize}$, the total time taken to balance the system. $t_{stabilize}$ is listed as t_stabilize in the chart below. The problem is to find a system design that minimizes $\phi(x, y) = t_{stabilize}$ subject to $F(x) = Y \wedge P(x, y)$ where:

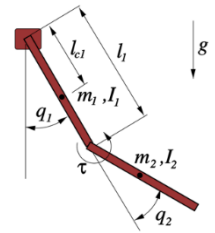

*Figure 18. Acrobot schematic*

$$P(x, y) = (I_1 = I_2 \wedge 0.1 \le m_1, m_2 \le 3.0 \wedge 0.1 \le I_1, I_2 \le 3.0 \wedge 0.1 \le l_1, l_2 \le 3.0 \wedge 0.3 \le l_{c_1}, l_{c_2} \le 0.7)$$

The constraint $I_1 = I_2$ reflects OpenAI/Gym's implementation of the Acrobot problem.

With 20 initial samples and a budget of 5000 seconds, CNMA finds solutions with objective function values 3.4, 3.2, and 2.8 with 1, 5, and 10 solvers, respectively. In the same amount of time, BO-S is able to find a solution with an objective function of 3.2 and NM-S is able to find a solution with an objective function of 8.6. Out of 500 randomly generated solutions, the best solution found has an objective function value of 4.2. Figure 19 and Figure 20 show the performance of CNMA, BO-S, NM-S, and Random Search with respect to time and number of function evaluations.

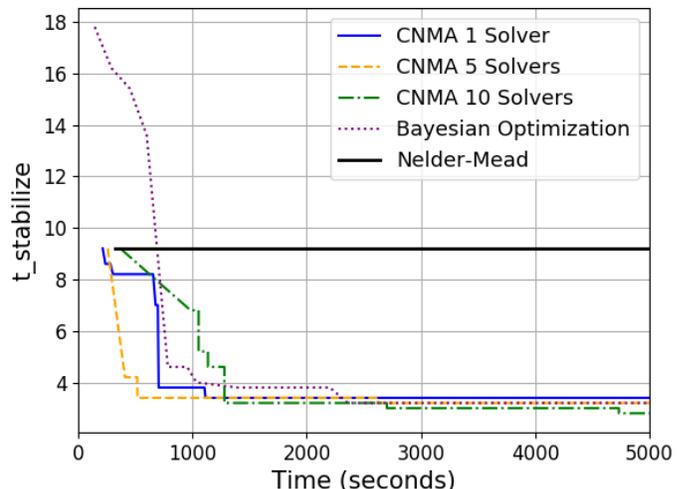

*Figure 19. Stabilization time vs. time for Acrobot. CNMA outperforms NM-S when given a time budget of less than 5000 seconds. CNMA matches or outperforms BO-S when given a time budget of less than 5000 seconds.*

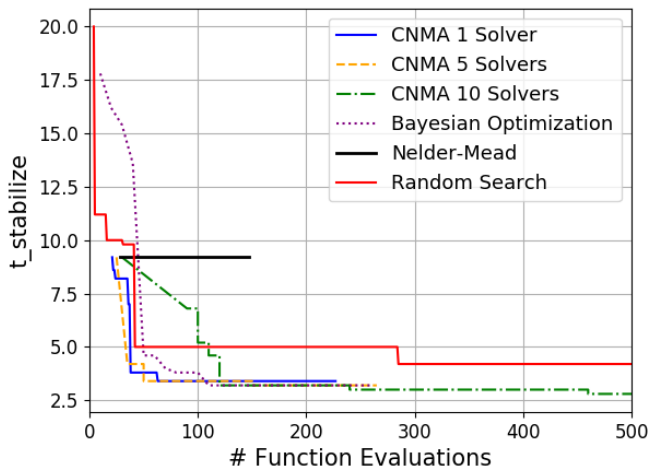

Figure 20. Stabilization time vs. number of function evaluations for Acrobot. CNMA matches or outperforms NM-S, BO-S, and Random Search when given the same budget of function evaluations.

### E. Optimal Sensor Placement for Power Grids

We now design a system in which all inputs are discrete but the output is continuous. Placing current and voltage sensors on a power grid can help identify which power lines are down during power outages. However, placing these sensors can be expensive and their number has to be limited. Given a limited sensor budget, the problem is to determine where to place the sensors such that their readings give the best chance of predicting the line failure pattern. We define a forward function $F$ that takes in a sensor placement $x$ of bits and outputs $y = [ambiguity]$, the "ambiguity" of the placement. In $x$, if a sensor is placed at line $i$, then $x[i] = 1$ else $x[i] = 0$. For a given $x$, we simulate all single line failures and record the associated set of readings at the sensors placed in $x$. If a reading set appears more than once then it is ambiguous since more than one failure can cause it. We then divide the number of ambiguous reading sets by the total number of reading sets to compute $[ambiguity]$ of $x$. Sensor readings are computed by the power line simulator OpenDSS [47], which takes in a model of the power grid, sensor placement, and a line failure pattern and outputs sensor readings. The objective function to minimize is $\phi(x, y) = ambiguity$ subject to the constraint that the number of sensors placed at most meets some budget. For the Bus118 power grid with 186 power lines, and a budget of 50 sensors, the constraint is specified as $P(x, y) = \text{sum}(x) \le 50$.

With 30 initial samples and a time budget of 1500 seconds, CNMA finds solutions with objective function values 0.04813, 0.02139, and 0.02139 with 1, 5, and 10 solvers, respectively. In the same amount of time, BO-S and NM-S are not able to find solutions that meet the constraints. Since these two only work with real-valued variables, we round their outputs to 1 or 0 to evaluate the forward function. Out of 460 randomly generated samples, none of them are valid solutions. Figure 21 and Figure 22 show the performance of CNMA, BO-S, NM-S and Random Search with respect to time and number of function evaluations.

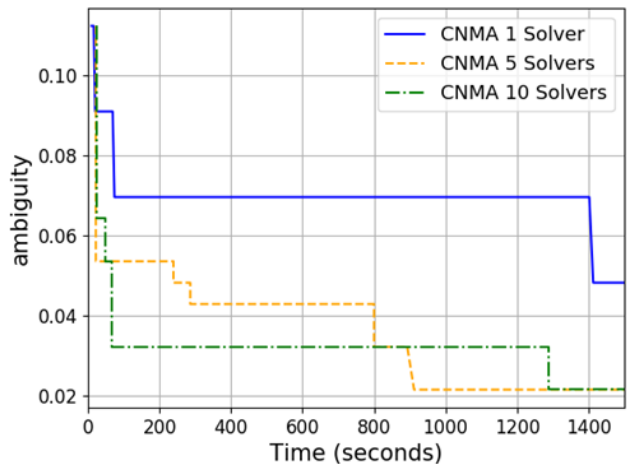

Figure 21. Ambiguity of the sensor placement vs. time for CNMA for sensor placement. All three CNMA runs find a solution with an ambiguity less than 0.05 within 1500 seconds. BO-S, NM-S and Random Search are not shown since these are not able to find any solutions.

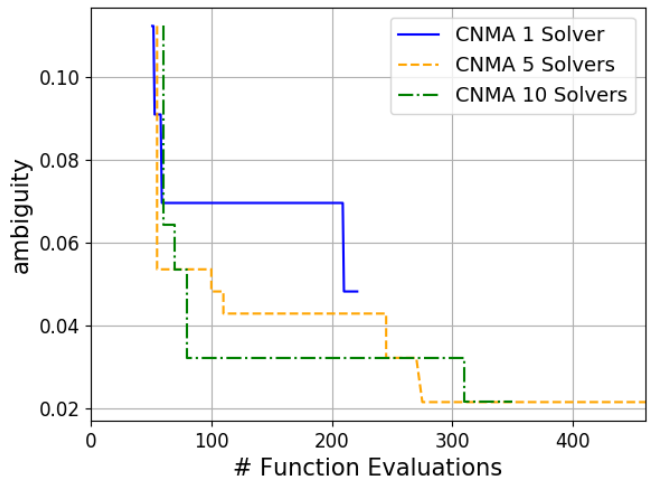

Figure 22. Ambiguity of the sensor placement vs. number of function evaluations for CNMA for sensor placement. All three CNMA runs find a solution with an ambiguity less than 0.05 with less than 300 function calls. BO-S, NM-S, and Random Search are not shown since they are not able to find any solutions.

### F. Rover Path Planning

This problem, defined in [20], involves finding a trajectory, such as the one shown in Figure 23, for a robot with starting and end goal positions. The trajectory is specified by a set of 30 2D points that are fit by a BSpline to define a path. Each trajectory also has an associated cost which penalizes obstacle collisions and should be minimized. The CNMA function $F$ takes as input the 2D coordinates of the 30 points (60 inputs) that define a path and outputs the cost of the trajectory. Each input ranges from 0.0 to 1.0. The objective is to maximize the negative cost of the trajectory.

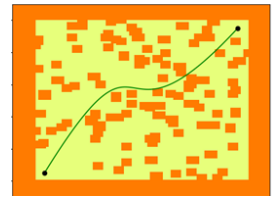

Figure 23. Path avoiding obstacles in the x-y plane

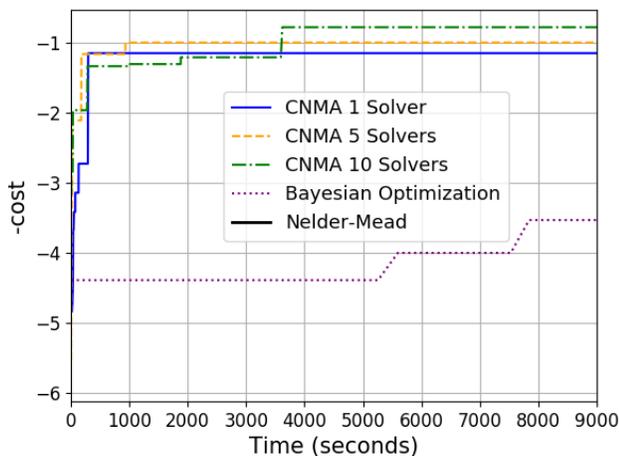

*Figure 24. Negative cost vs. time for Rover Path Planning. CNMA outperforms BO-S and NM-S when given a time budget of less than 9000 seconds. Note that NM-S stops after 6 seconds due to early convergence.*

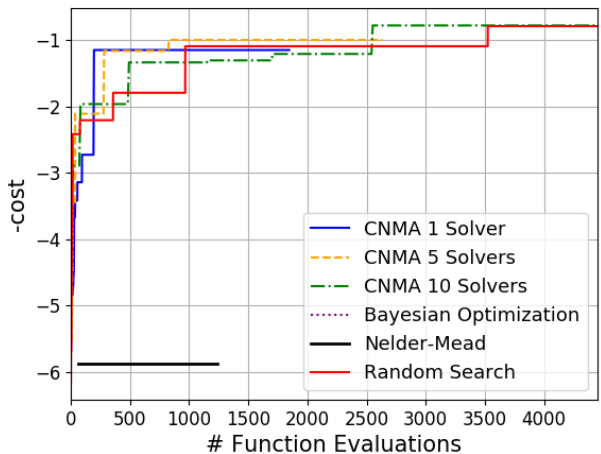

*Figure 25. Negative cost vs. number of function evaluations for Rover Path Planning. CNMA outperforms both NM-S and BO-S. While Random Search and CNMA with 10 solvers find similar solutions, CNMA with 10 solvers finds the solution in fewer function evaluations. Note that BO-S is only able to complete 30 function evaluations in the given time budget.*

With two initial samples and a time budget of 9000 seconds, CNMA finds solutions with costs of 1.148, 0.997, and 0.778 with 1, 5, and 10 solvers, respectively. Given the same time budget, BO-S finds a solution with a cost of 3.530 and NM-S finds a solution with a cost of 5.867. Out of 4452 random samples, the best solution has a cost of 0.788. Figure 24 and Figure 25 shows the performance of CNMA, BO-S, NM-S, and Random Search with respect to time and number of function evaluations.

### G. Polak3

This optimization benchmark [48] involves minimizing the maximum value of 10 different transcendental functions. The CNMA forward function $F(x) = y$ takes in 10 values, $x_1, ..., x_{10}$, each $x_i \in [-1, 1]$, and outputs $y$, the maximum value of the 10 transcendental functions. The best known minimum is 5.93. An example of one of the transcendental functions is $\left(e^{(x_1 - \sin(0.0 + 1.0 + 1.0)) * (x_1 - \sin(0.0 + 1.0 + 1.0))}\right) +$
$0.5 * \left(e^{(x_2 - \sin(0.0 + 2.0 + 2.0)) * (x_2 - \sin(0.0 + 2.0 + 2.0))}\right) +$

$0.3333 * \left(e^{(x_3 - \sin(0.0 + 3.0 + 3.0)) * (x_3 - \sin(0.0 + 3.0 + 3.0))}\right) +$
$0.25 * \left(e^{(x_4 - \sin(0.0 + 4.0 + 4.0)) * (x_4 - \sin(0.0 + 4.0 + 4.0))}\right) +$
$0.2 * \left(e^{(x_5 - np.\sin(0.0 + 5.0 + 5.0)) * (x_5 - \sin(0.0 + 5.0 + 5.0))}\right) +$
$0.1666 * \left(e^{(x_6 - \sin(0.0 + 6.0 + 6.0)) * (x_6 - \sin(0.0 + 6.0 + 6.0))}\right) +$
$0.1428 * \left(e^{(x_7 - \sin(0.0 + 7.0 + 7.0)) * (x_7 - \sin(0.0 + 7.0 + 7.0))}\right) +$
$+ 0.125 * \left(e^{(x_8 - \sin(0.0 + 8.0 + 8.0)) * (x_8 - \sin(0.0 + 8.0 + 8.0))}\right) +$
$0.1111 * \left(e^{(x_9 - \sin(0.0 + 9.0 + 9.0)) * (x_9 - \sin(0.0 + 9.0 + 9.0))}\right) +$
$0.1 * \left(e^{(x_{10} - \sin(0.0 + 10.0 + 10.0)) * (x_{10} - \sin(0.0 + 10.0 + 10.0))}\right) +$
$0.0909 * \left(e^{(x_{11} - \sin(0.0 + 11.0 + 11.0)) * (x_{11} - \sin(0.0 + 11.0 + 11.0))}\right)$

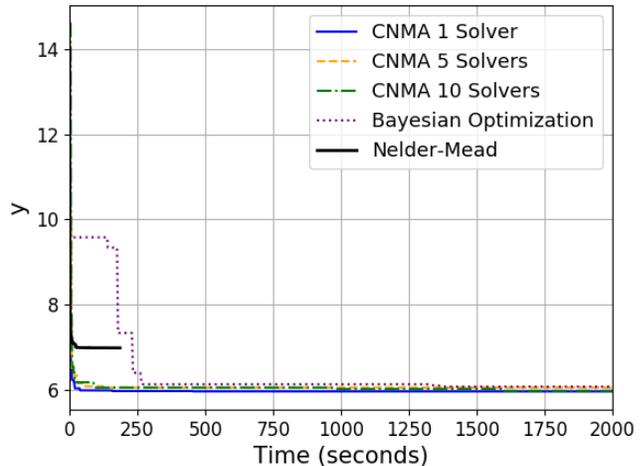

*Figure 26. The maximum value of the 10 transcendental functions vs. time Polak3. While CNMA and BO-S are able to find solutions with similar objective values, it takes BO-S approximately 250 seconds longer. Note that NM-S stops after 186 seconds due to early convergence.*

With 20 initial samples and a time budget of 2000 seconds, with 1, 5, and 10 solvers CNMA finds solutions with objective function values 5.97, 6.06, and 5.98, respectively. Given the same time budget, BO-S finds a solution with an objective function value of 6.08 and NM-S finds a solution with an objective function value of 6.99. Out of 2680 randomly generated samples, the best solution has an objective function of 6.81. Figure 26 and Figure 27 show the performance of CNMA, BO-S, NM-S, and Random Search with respect to time and number of function evaluations.

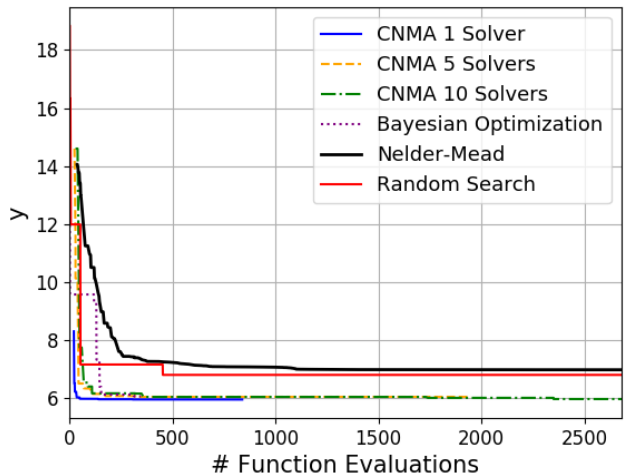

*Figure 27: The maximum value of the 10 transcendental functions vs. number of function evaluations for Polak3. Random Search and NM-S find solutions with an objective around 6.9 while CNMA and BO-S are able to find better solutions with an objective around 6.*

### H. Modeling Nonlinear Constraints and Objective Functions

To model a nonlinear constraint $P$, we add, for each nonlinear expression in $P$, an extra output to $F$ denoting the value of the expression and then express $P$ as a linear constraint on these outputs. For example, we show how to solve a benchmark problem in [19]: minimize $g(x_1, x_2)$ subject to $c_1(x_1, x_2) \geq 0 \land c_2(x_1, x_2) \geq 0$ where $g(x_1, x_2) = x_1 + x_2$, $c_1(x_1, x_2) = 0.5 * \sin(2\pi (x_1^2 - 2x_2)) + x_1 + 2x_2 + 1.5$, $c_2(x_1, x_2) = -(x_1^2) - (x_2^2) + 1.5$. We define a function $F(x_1, x_2)$ that produces two outputs $v_1, v_2$ computing, respectively, $c_1(x_1, x_2)$ and $c_2(x_1, x_2)$. Then with CNMA we solve the problem of minimizing $x_1 + x_2$ subject to $F(x_1, x_2) = [v_1, v_2] \land v_1 \geq 0, v_2 \geq 0$. In 84 evaluations (10 initial + 74 additional), CNMA finds a solution of 0.6003. The minimum is 0.599. Our neural network architecture has 35 and 10 neurons in the two hidden layers. The important point to note is that we can solve this problem through specification, not by changing CNMA. If the objective function $g$ had been nonlinear, we would have added an extra output to $F$ and minimized that.

## VI. CONCLUSIONS

System design tools are often only available as blackboxes with complex nonlinear relationships between inputs and outputs. This article presents CNMA, a new constrained optimization method for blackboxes for solving the inverse problem of finding designs from requirements on output.

CNMA's innovation is connecting the modeling power of neural networks and constraint-solving power of MILP solvers into a learning-from-failure feedback loop in such a way that they do much of the work for us, permitting straightforward implementations of several desirable features into a single, cohesive system: efficient surrogate function construction, sample efficiency, constraint solving without penalty functions, solving blackbox constraints, optimization with discrete and continuous variables, resilience to nonterminating function evaluations, and parallelism.

If a large and deep neural network is needed to model a complex function, its MILP equivalent may not be efficiently solvable. However, CNMA does not need to model the function in its entire domain. It only needs to model it in the part of the domain relevant to solving the constrained optimization problem. If this region is not too complex, a smaller neural network is adequate so that its MILP equivalent could be efficiently solvable. This region is automatically computed by CNMA. As we have seen, the largest network used had [35, 10] neurons in its hidden layers and most problems start with few tens of initial samples, some even with just two. In fact, a large or deep neural network may be detrimental to performance as it would overfit the small number of points CNMA samples.

CNMA is evaluated for seven nonlinear design problems of 8 (2 problems), 10, 15, 36 and 60 real-valued dimensions and one with 186 binary dimension. It is shown that CNMA improves upon stable, off-the-shelf implementations of BO/GP (BO-S), Nelder Mead (NM-S), and Random Search by 1%-87% for a fixed time and function evaluation budget. Note, however, that BO-S did not return a solution for two problems and NM-S did not return one for three. Future research problems include introducing additional diversity, e.g., via bootstrapping, multi-function CNMA, and finding a good initial neural network architecture and adapting it as new samples are created.

**Acknowledgments**. We thank the anonymous reviewers and Antoine Cully, Robert Gramacy, Matthias Poloczek, Robert Vanderbei, John Chinneck, Nick Sahinidis, Jaime Fisac, Nicholas Kraus, Taylor Njaka, Jeremy Cohen, Todd Huster, Mengdi Wang and Chetan Narain, for helpful comments.

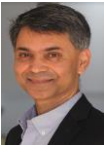
**Sanjai Narain** is Fellow and Chief Scientist at Peraton Labs, Basking Ridge, NJ. He received his B.Tech. in EE from IIT Delhi in 1979 and his Ph.D. in CS from UCLA in 1988.

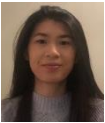
**Emily Mak** is Research Scientist at Peraton Labs, Basking Ridge, NJ. She received her B.S. and M.S. in Applied Math from Johns Hopkins University in 2018.

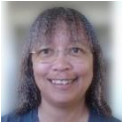
**Dana Chee** is Senior Research Scientist at Peraton Labs, Basking Ridge, NJ. She received her M.S. in Electrical Engineering from Howard University.

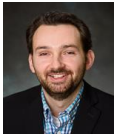
**Brendan Englot** is Geoffrey Inman Professor of Mechanical Engineering at Stevens Institute of Technology, Hoboken, NJ. He received his B.S., M.S. and Ph.D. in Mechanical Engineering from MIT in 2007, 2009, and 2012, respectively.

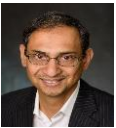
**Kishore Pochiraju** is Professor of Mechanical Engineering at Stevens Institute of Technology, Hoboken, NJ. He received his M. Tech. in Mechanical Engineering from IIT Kanpur and a Ph.D. from Drexel University in 1993.

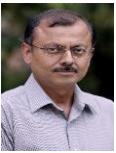
**Niraj K. Jha** is Professor of Electrical Engineering at Princeton University, Princeton, NJ. He obtained his B.Tech. in E&ECE from IIT Kharagpur in 1981 and a Ph.D. in EE from University of Illinois at Urbana-Champaign, IL in 1985. He is a Fellow of IEEE and ACM.

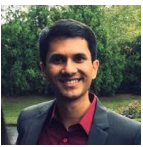
**Karthik Narayan** is the CEO of Starfruit-LLC. He obtained his B.S. in Math/CS from Georgia Tech. in 2011 and a Ph.D. in CS/AI from University of California, Berkeley in 2016.